# Layout Control and Semantic Guidance with Attention Loss Backward for T2I Diffusion Model

Guandong Li[1*]

1.*Suning, Xuanwu, Nanjing, 210042, Jiangsu, China.

*Corresponding author(s). E-mail(s): leeguandon@gmail.com

***Abstract***: Controllable image generation has always been one of the core demands in image generation, aiming to create images that are both creative and logical while satisfying additional specified conditions. In the post-AIGC era, controllable generation relies on diffusion models and is accomplished by maintaining certain components or introducing inference interferences. This paper addresses key challenges in controllable generation: 1. mismatched object attributes during generation and poor prompt-following effects; 2. inadequate completion of controllable layouts. We propose a train-free method based on attention loss backward, cleverly controlling the cross attention map. By utilizing external conditions such as prompts that can reasonably map onto the attention map, we can control image generation without any training or fine-tuning. This method addresses issues like attribute mismatch and poor prompt-following while introducing explicit layout constraints for controllable image generation. Our approach has achieved excellent practical applications in production, and we hope it can serve as an inspiring technical report in this field.

1. Introduction

   Controllable generation based on diffusion models has already yielded significant benefits in the field of creative generation. Controllable creative image generation is also one of the core tasks in the e-commerce sector, which refers to T2I generation that retains a certain level of imagination and creativity while meeting pre-specified physical attributes, such as the number, position, size, and material of objects. In the pre-AIGC era, e-commerce controllable generation typically relied on material parsing[1], synthesis[2], and deployment[3] to obtain good, dynamically updated constrained creative templates. Specific product and copy area layouts would then be arranged based on these templates, constraining the generated creativity within a fixed space. With the emergence of stable diffusion[4], SD can generate images with excellent style and lighting, but this is not sufficient. Therefore, there are numerous methods to optimize controllable SD generation. This paper focuses on controllable adjustments in image generation based on layout control and semantic guidance dimensions.

   Controllable generation is generally divided into two main types. The first type focuses on interference reasoning path generation, which involves influencing the final generation results through methods such as concatenating attention layers or noise during the generation process. A typical example is prompt2prompt[5], where the cross-attention corresponding to text tokens contains strong semantic information, revealing how the attention mechanism drives T2I tasks. This method allows for controllable image generation by modifying the token embeddings of text in attention. The second approach enhances the model through fine-tuning or the addition of new network layers, represented by ControlNet[6], which requires fine-tuning the SD model to generate images under certain constraint conditions. The first approach can be further divided into two technical routes: forward guidance and backward guidance, where forward guidance involves applying regional restrictions during the inference process, and backward guidance is similar to gradient update methods for model optimization. The method proposed in this paper is based on the first approach and is train-free.

In controllable image generation, especially in the e-commerce sector[7], several core issues typically need to be addressed: 1. The problem of attribute mismatch at the semantic level, where the model binds attributes to the wrong subjects or fails to bind them entirely, resulting in images that semantically do not meet the requirements of the prompt; 2. The deficiencies of using prompts for description, which lack explicit layout input control, making it impossible to capture the layout of image generation when describing certain spatial relationships. To address issue 1, we propose semantic guidance, which leverages cross-attention map information to adjust the intermediate latent during the denoising process, enabling a stronger mapping between the text prompt and the activation values in the activation map, thereby guiding the model to generate all subjects described in the text prompt. To tackle issue 2, we explicitly introduce layout information, sampling from an additional controlled distribution to guide the layout during the generation process. The user-specified layout corresponds to the selected text tokens, allowing spatial layout adjustments of the generated images through cross-attention.

We are based on a train-free approach that does not require training or fine-tuning, and we have adopted a loss backward guidance method to impose loss constraints on the attention map, enabling the model to generate controllable images. Our contributions include: 1. simultaneously improving the methods for addressing attribute mismatch and layout optimization, enhancing controllable image generation; 2. utilizing a train-free approach and employing attention loss backward to accomplish controllable image generation.

2. Related works

2.1 Layout control

ReCo[8] introduces target boxes as additional input conditions, concatenating the corresponding coordinates with text descriptions to form more specific prompts that drive region-controlled generation. However, this approach requires fine-tuning based on the SD model. GLIGEN[9] also aims to control the layout of generated images using a set of text and a set of target box inputs but employs a more flexible and lightweight design. ReCo integrates the features of positional coordinates and text descriptions while keeping the main network structure unchanged, which necessitates fine-tuning the CLIP text encoder and Unet to accommodate changes in data flow. In contrast, GLIGEN adds more new layers to handle these additional inputs, allowing for the original network to be frozen to save computational resources. By intervening in cross-attention and using user-specified inputs to guide generation in the chosen direction, backward guidance is employed to update the image's latent through backpropagation, minimizing energy to match the desired layout[10].

2.2 Segmantic guidance

Semantic guidance is often processed based on the qkv of the attention layer. Qkv refers to the module used to align and fuse modalities such as text and image into the generative model, typically found in the cross-attention and self-attention mechanisms of Unet. For example, the paper "Condition and Mask Guidance"[11] processes the attention map formed by qk through text prompt guidance and attention mask guidance. The concept of Generative Semantic Nursing is introduced in "Attend-and-Excite,"[12] where we intervene in the generation process in real-time during inference to enhance the credibility of the generated images. In Prompt2Prompt, different text features from CLIP are mapped as values, which are continuously reused and stacked at different positions based on attention weights, thereby controlling image generation by modifying the cross-attention values.

3. Method

This section will provide a detailed introduction to our controllable image generation framework. In our train-free approach, we utilize the attention loss backward method to optimize the final image generation. Among several effective means of controllable generation, prompt input and explicit input are two very commonly used methods. We combine these two approaches to manipulate the attention map based on the consistency between the attention map and the final generated image in the cross-attention dimension.

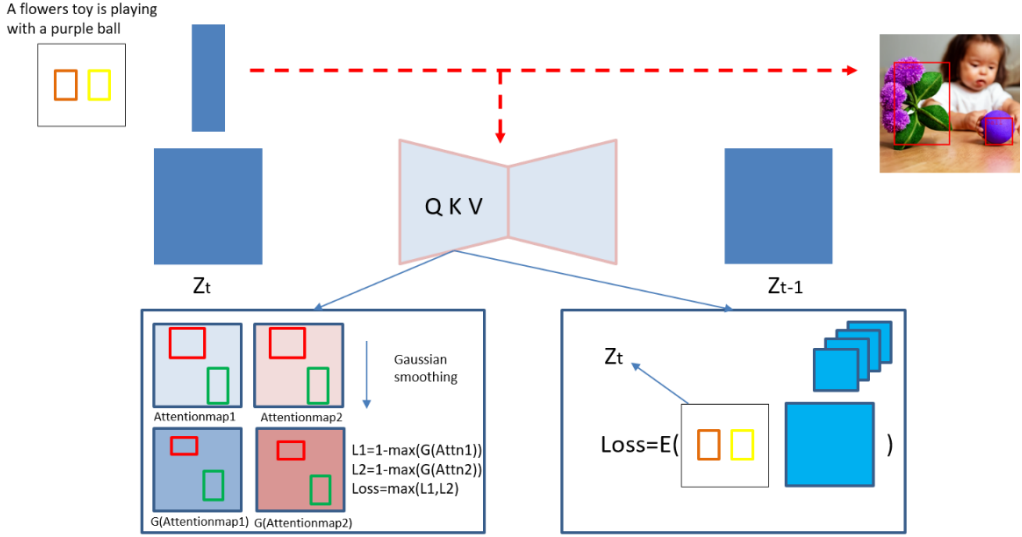

Figure 1: controllable image generation framework

3.1 Segmantic guidance

Using masks or semantic guidance to direct the attention map is an effective method. For each subject token S in the prompt, there should be at least one patch in the attention map with a high attention value corresponding to it. To strengthen the mapping relationship between the attention map and the token, Gaussian filtering can be applied to the attention map. For the subject token that is most disregarded at the current time step, the optimized loss function can be defined as follows:

$$L = \max L_s \text{ where } L_s = 1 - \max(attention\ map)$$

Because different subject tokens may be activated at different time steps, all the disregarded subject tokens may be activated at some point in time.

After calculating the loss function L, the current latent Zt will be updated using gradients:

$$Z'_t = Z_t - \alpha_t * \nabla_{zt} L$$

Here, $\alpha_t$ defines the gradient update step size, and this step is similar to the update steps in classifier-guided diffusion.

Finally, use $Z'_t$ in place of $Z_t$ to continue the denoising sampling process. This gradient update should not be applied to all time steps but rather only to a subset of time steps t = (T, T-1, …, $T_{end}$) \). According to the default setting in SD, T = 50, and based on the final experimental observations, $T_{end} = 25$ is determined, at which point the spatial positions of objects in the generated images are not altered, resulting in favorable outcomes. If the attention value of a subject token does not reach a certain threshold during the early stages of denoising, the corresponding subject will not be generated. Therefore, the gradient update process should be revised to iteratively update $Z_t$ until all subject tokens reach the predefined minimum attention value.

3.2 Layout control

When we input a layout, we are essentially providing a set of coordinates. The reverse guidance defines a function to evaluate the degree of aggregation of the specified token's cross-attention within the designated box B:

$$E(A, B, i) = \left(1 - \frac{\sum_{p \in B} Attentionmap}{\sum_p Attentionmap}\right)^2$$

And the gradient of this metric is updated to the latent at the current time step. Optimizing this function can result in a higher value for the cross-attention map of the i-th token within the area specified by B, thereby guiding the image generation according to the layout.

4. Results

As shown in the figure below, we demonstrate how we address attribute mismatches and introduce layouts to solve layout issues, highlighting the visual effects of the generated images.

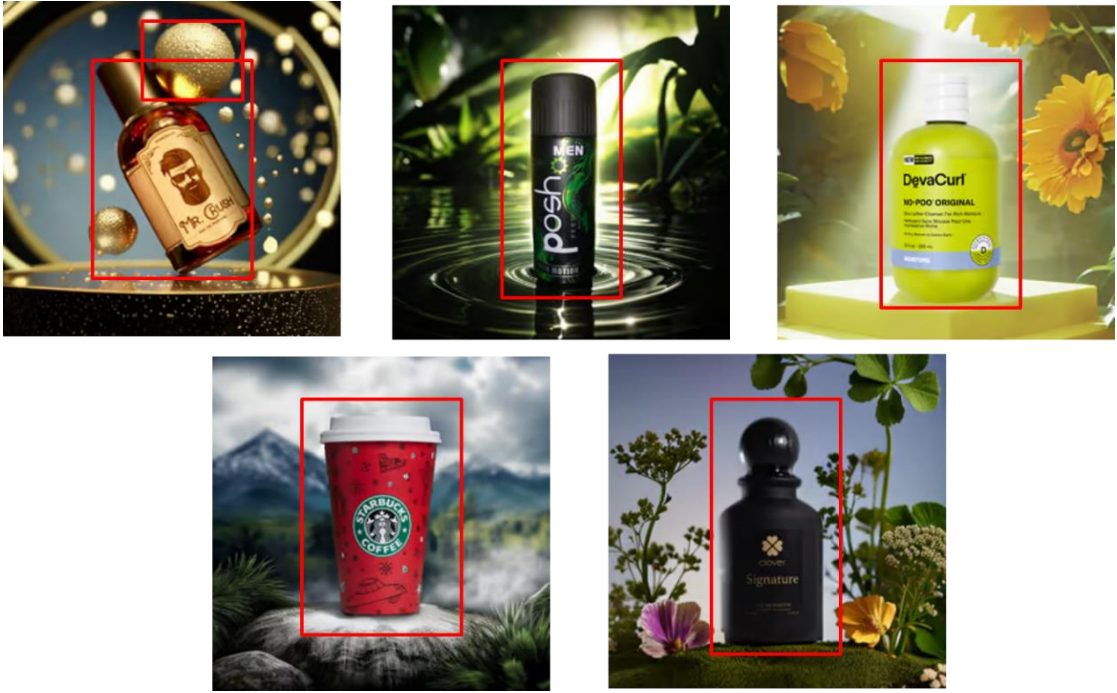

Figure 2: Model results

5. conclusion

This paper presents a comprehensive method for addressing core issues in controllable image generation, specifically focusing on attribute mismatches and the lack of layout introduction. The framework proposed is train-free and designed around two external conditions: prompts and layout information. We developed an attention loss backward method that cleverly utilizes the consistency and controllability of the attention map with respect to these external conditions, applying constraints on the attention map to resolve the issues of controllable generation. This provides effective solutions for both prompt-following and layout-following in practical applications. We hope this paper can serve as an inspiring technical report in the field of controllable image generation.

References

1. Li G, Yang X. Intelligent Parsing: An Automated Parsing Framework for Extracting Design Semantics from E-commerce Creatives[J]. arXiv preprint arXiv:2312.17283, 2023.
2. Li G, Yang X. Smartbanner: intelligent banner design framework that strikes a balance between creative freedom and design rules[J]. Multimedia Tools and Applications, 2023, 82(12): 18653-18667.
3. Li G, Yang X. Two-stage dynamic creative optimization under sparse ambiguous samples for e-commerce advertising[J]. arXiv preprint arXiv:2312.01295, 2023.
4. Rombach R, Blattmann A, Lorenz D, et al. High-resolution image synthesis with latent diffusion models[C]//Proceedings of the IEEE/CVF conference on computer vision and pattern recognition. 2022: 10684-10695.
5. Hertz A, Mokady R, Tenenbaum J, et al. Prompt-to-prompt image editing with cross attention control[J]. arXiv preprint arXiv:2208.01626, 2022.
6. Zhang L, Rao A, Agrawala M. Adding conditional control to text-to-image diffusion models[C]//Proceedings of the IEEE/CVF International Conference on Computer Vision. 2023: 3836-3847.
7. Li G. E-Commerce Inpainting with Mask Guidance in Controlnet for Reducing Overcompletion[J]. arXiv preprint arXiv:2409.09681, 2024.
8. Yang Z, Wang J, Gan Z, et al. Reco: Region-controlled text-to-image generation[C]//Proceedings of the IEEE/CVF Conference on Computer Vision and Pattern Recognition. 2023: 14246-14255.
9. Li Y, Liu H, Wu Q, et al. Gligen: Open-set grounded text-to-image generation[C]//Proceedings of the IEEE/CVF Conference on Computer Vision and Pattern Recognition. 2023: 22511-22521.
10. Chen M, Laina I, Vedaldi A. Training-free layout control with cross-attention guidance[C]//Proceedings of the IEEE/CVF Winter Conference on Applications of Computer Vision. 2024: 5343-5353.
11. Li G. Training-Free Style Consistent Image Synthesis with Condition and Mask Guidance in E-Commerce[J]. arXiv preprint arXiv:2409.04750, 2024.
12. Chefer H, Alaluf Y, Vinker Y, et al. Attend-and-excite: Attention-based semantic guidance for text-to-image diffusion models[J]. ACM Transactions on Graphics (TOG), 2023, 42(4): 1-10.